%% file: main.tex
\documentclass[10pt,twocolumn,letterpaper]{article}

\usepackage{iccv}              

\input{preamble}

\input{color_for_revision}

\title{\method: Mitigating Video Hallucinations by Prompt-Aware \\ Multi-Instance Video Preference Learning}

\author{%
 Xinpeng Ding$^{1}$,
  Kui Zhang$^{2}$,
  Jianhua Han$^{2}$,
  Lanqing Hong$^{2}$,
  Hang Xu$^{2}$,
  Xiaomeng Li$^{1}$\\
$^{1}$The Hong Kong University of Science and Technology \\
\enspace
$^{2}$Huawei Noah's Ark Lab\\
\vspace{-0.5cm}
}

\begin{document}
\maketitle
\input{ICCV2025-Author-Kit-Feb/sec/0_abstract}    
\input{ICCV2025-Author-Kit-Feb/sec/1_intro}

\input{ICCV2025-Author-Kit-Feb/sec/2_related_work}

\input{ICCV2025-Author-Kit-Feb/sec/3_preliminary}

\input{ICCV2025-Author-Kit-Feb/sec/4_vdpo}

\input{ICCV2025-Author-Kit-Feb/sec/5_method}

\input{ICCV2025-Author-Kit-Feb/sec/6_experiments}

\input{ICCV2025-Author-Kit-Feb/sec/7_conclusion}

{
    \small
    \bibliographystyle{ieeenat_fullname}
    \bibliography{main}
}

\input{supply}

\end{document}

%% file: preamble.tex
\definecolor{iccvblue}{rgb}{0.21,0.49,0.74}
\PassOptionsToPackage{table}{xcolor}
\usepackage[pagebackref,breaklinks,colorlinks,allcolors=iccvblue]{hyperref}
\usepackage{graphicx}
\usepackage{xcolor}         
\usepackage{amsmath}

\usepackage{multirow}
\usepackage{enumitem}
\usepackage{booktabs}       
\usepackage{amsfonts}       
\usepackage{nicefrac}       
\usepackage{microtype}      
\usepackage{adjustbox}
\usepackage{xspace}    
\usepackage{colortbl}
\usepackage{pifont}
\usepackage{wrapfig}
\usepackage[table]{xcolor}
\usepackage{enumitem}
\usepackage{transparent}
\usepackage{algorithm}
\usepackage{algorithmic}

\usepackage{amssymb}
\usepackage{mathtools}
\usepackage{booktabs}
\usepackage{tabularx}
\usepackage{colortbl}
\usepackage{bbding}
\usepackage{makecell}
\usepackage{bbding}
\usepackage{amssymb}
\usepackage{pifont}
\usepackage[accsupp]{axessibility}
\usepackage{textcomp}
\usepackage{tikz}
\newlength\savewidth\newcommand\shline{\noalign{\global\savewidth\arrayrulewidth
  \global\arrayrulewidth 1pt}\hline\noalign{\global\arrayrulewidth\savewidth}}
\newcommand{\tablestyle}[2]{\setlength{\tabcolsep}{#1}\renewcommand{\arraystretch}{#2}\centering\footnotesize}
\renewcommand{\paragraph}[1]{\vspace{0.1mm}\noindent\textbf{#1}}
\newcommand{\baseline}[1]{\cellcolor{baselinecolor}{#1}}
 
\newcolumntype{x}[1]{>{\centering\arraybackslash}p{#1pt}}
\newcolumntype{y}[1]{>{\raggedright\arraybackslash}p{#1pt}}
\newcolumntype{z}[1]{>{\raggedleft\arraybackslash}p{#1pt}}

\newcommand{\app}{\raise.17ex\hbox{$\scriptstyle\sim$}}

\definecolor{deemph}{gray}{0.6}

\definecolor{baselinecolor}{gray}{.9}

\newcommand{\method}{{PaMi-VDPO}}

\newcommand{\sm}{supplementary materials}
\newcommand{\weight}{question-aware weight}

\definecolor{pearDark}{HTML}{2980B9}
\definecolor{tabfirstred}{rgb}{1, 0.7, 0.7}
\colorlet{tabfirst}{pearDark}

\definecolor{light-gray}{gray}{0.6}
\definecolor{front-color}{HTML}{F5FFFA}
\definecolor{tabhighlight}{HTML}{e5e5e5}
\definecolor{improvement}{RGB}{225,97,78}
\definecolor{lightblue}{HTML}{B3E5FC}
\definecolor{mygreen}{rgb}{0.0, 0.51, 0.0}
\definecolor{myblue}{rgb}{0.0, 0.0, 1.5}
\definecolor{mygray}{gray}{0.93}

\definecolor{tabsecondcolor}{rgb}{1, 0.85, 0.7}


%% file: color_for_revision.tex
\usepackage{color}
\definecolor{green}{rgb}{0, 0.5, 0}
\definecolor{orange}{rgb}{0.8, 0.6, 0.2}
\definecolor{red}{rgb}{1.0, 0.0, 0.0}
\definecolor{teal}{rgb}{0.0, 0.4, 0.4}
\definecolor{purple}{rgb}{0.65,0,0.65}
\definecolor{saffron}{rgb}{0.95,0.75,0.2}
\definecolor{turquoise}{rgb}{0.0,0.5,0.5}
\definecolor{black}{rgb}{0.0, 0.0, 0.0}
\definecolor{gray}{rgb}{0.5, 0.5, 0.5}

%% file: ICCV2025-Author-Kit-Feb/sec/0_abstract.tex
\begin{abstract}
Direct Preference Optimization (DPO) helps reduce hallucinations in Video Multimodal Large Language Models (VLLMs), but its reliance on offline preference data limits adaptability and fails to capture true video-response misalignment.
We propose Video Direct Preference Optimization (VDPO), an online preference learning framework that eliminates the need for preference annotation by leveraging video augmentations to generate rejected samples while keeping responses fixed. However, selecting effective augmentations is non-trivial, as some clips may be semantically identical to the original under specific prompts, leading to false rejections and disrupting alignment. To address this, we introduce \textbf{P}rompt-\textbf{a}ware \textbf{M}ulti-\textbf{i}nstance Learning VDPO (\method), which selects augmentations based on prompt context. Instead of a single rejection, we construct a candidate set of augmented clips and apply a close-to-far selection strategy, initially ensuring all clips are semantically relevant while then prioritizing the most prompt-aware distinct clip. This allows the model to better capture meaningful visual differences, mitigating hallucinations, while avoiding false rejections, and improving alignment. \method~seamlessly integrates into existing VLLMs without additional parameters, GPT-4/human supervision. With only 10k SFT data, it improves the base model by 5.3\% on VideoHallucer, surpassing GPT-4o, while maintaining stable performance on general video benchmarks.
\end{abstract}

%% file: ICCV2025-Author-Kit-Feb/sec/1_intro.tex
\input{ICCV2025-Author-Kit-Feb/items/fig_intro_v2}

\section{Introduction}\label{sec:intro}

Video multi-modal large language models (VLLMs)~\cite{liu2023visual,zhu2023minigpt,liu2024improved,li2023llava,li2023blip} have demonstrated remarkable capabilities in video understanding tasks, such as video question answering and captioning~\cite{zhang2024llavanextvideo,li2023videochat,lin2023video,maaz2023video,ding2023hilm,ding2024holistic}. However, these models suffer from a persistent challenge: {hallucination}~\cite{liu2024survey}, where generated responses deviate from actual visual content. While hallucination in image-based MLLMs has been widely studied~\cite{yin2024woodpecker,lee2024volcano,zhou2023analyzing,leng2024mitigating,zhao2024mitigating,yu2024hallucidoctor,li2023silkie,fu2025chip,xie2024v,wang2024mdpo,yu2024rlaifv}, its impact and mitigation in VLLMs remain largely unexplored.

Existing efforts to mitigate hallucination in VLLMs fall into two main categories. \textbf{(i) Architectural modifications}, such as VISTA-LLAMA~\cite{ma2024vista}, which alter attention mechanisms to reduce hallucinations. However, these modifications require retraining for different tasks and hinder the use of efficient implementations like FlashAttention~\cite{dao2022flashattention}. \textbf{(ii) Preference learning}, particularly {Direct Preference Optimization (DPO)}~\cite{zhang2024direct}, which trains models to prefer more reliable responses over hallucinated ones based on the constructed video preference data.

While effective, DPO has several limitations, as shown in Fig.\ref{fig:intro}(a):
\textbf{(i) High cost of pre-construction.} DPO relies on human annotators or powerful LLMs to generate high-quality preference data, making it costly and labor-intensive.
\textbf{(ii) Offline limitation.} Preference data is generated before training, leading to two major drawbacks: \textbf{lack of adaptability}, as offline data cannot adjust to new tasks or scenarios, and \textbf{data staleness}, where pre-constructed data may become outdated and misaligned with evolving models and task requirements.
\textbf{(iii) Response-focused preference.} Since optimization primarily targets response alignment, the model may prioritize linguistic style over addressing misalignment with visual content~\cite{wang2024mdpo} (see in Fig.~\ref{fig:visual}).

To address these limitations, this paper introduces a novel attempt, \textbf{Video DPO (VDPO)}, a video-centric online preference learning that obviates the requirement for pre-constructed preference data. Rather than annotating text responses, VDPO applies \emph{video augmentations to generate rejected video clips while sharing the original response}. By training the model to prefer the original video over its augmented variants, VDPO explicitly enforces video-response alignment, mitigating hallucinations based on visual content rather than textual style. Unlike traditional DPO, VDPO is entirely self-supervised, making it more scalable and adaptable to various video tasks.

\input{ICCV2025-Author-Kit-Feb/items/fig_intro_compare}

Through our experiments in Section~\ref{sec:bottleneck}, we identify \emph{augmentation selection as a {crucial} bottleneck in VDPO}, significantly impacting its effectiveness. 
A fundamental challenge lies in the trade-off between \textbf{general semantic similarity} and \textbf{prompt\footnote{"Prompt" is also referred to as "question" or "instruction."}-based distinctiveness}.
On one hand, the augmented rejected clips should maintain high semantic similarity with the original to help the model detect subtle inconsistencies for mitigating hallucinations (Fig.~\ref{fig:motivation}(b)). On the other hand, overly similar rejected clips can lead to false-rejected samples,~\ie, where the rejected clips are semantically identical to the original clip under specific prompts (Fig.~\ref{fig:analysis}).
This false-rejected issue would introduce alignment noise, leading to unstable training and suboptimal performance, as shown in Fig.\ref{fig:intro_compare} (RA).

To address this, we propose \textbf{P}rompt-\textbf{a}ware \textbf{M}ulti-\textbf{i}nstance Learning VDPO (\textbf{\method}), a novel framework that \emph{automatically constructs multiple candidates, allowing our model to dynamically adjust its preference learning based on prompts while mitigating interference from false-rejected noise.}
\method~is performed in a \textbf{close-to-far strategy}: We first generate a candidate set of rejected clips using high-similarity augmentations, ensuring that rejected samples remain visually close to the original video while capturing subtle hallucinations. Then, we select the most divergent augmented clip—i.e., the one whose response deviates most from the original—as the pseudo true-rejected sample. By dynamically prioritizing the most misaligned sample while down-weighting others, \method~seamlessly integrates prompt-dependent augmentation selection into preference learning through a multi-instance learning (MIL) approach. Notably, \method~requires no additional parameters, no architectural modifications, and no manually curated preference data, making it efficient and easily integrated into existing VLLM training pipelines.

\noindent\textbf{Contributions.} Our main contributions are:
\begin{itemize}
    \item We introduce {Video DPO (VDPO)}, the first video-centric {online preference optimization} framework that directly enforces video-response alignment through video augmentations, eliminating the reliance on pre-constructed preference data.
    \item We identify false rejection in video augmentations as a critical bottleneck in VDPO, where rejected clips remain overly semantic-similar to the original under the given prompt, introducing alignment noise.
    \item We propose {\method}, allowing the model to leverage the prompt information to adaptively learn from diverse augmented clips while suppressing noise ones to avoid false-rejected issues, improving stability and effectiveness.
    \item Our approach achieves {state-of-the-art hallucination mitigation} while ensuring stable general performance, without requiring additional preference data, model parameters, or architectural {modifications}.
\end{itemize}

%% file: ICCV2025-Author-Kit-Feb/items/fig_intro_v2.tex
\begin{figure}[t]
\centering 
\includegraphics[width=0.48\textwidth,height=0.2\textheight]{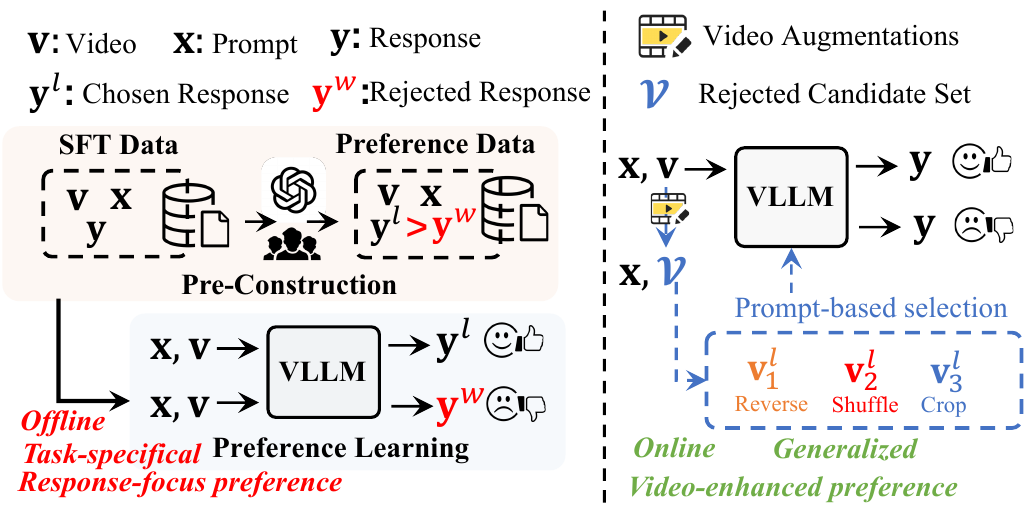}
\vspace{-0.8cm}
\caption{\textbf{Comparison of different methods.}  
\textbf{Left:} Direct Preference Optimization (DPO) requires costly LLMs or human annotators to preconstruct preference data offline for different tasks, limiting its generalization. Meanwhile, its preference learning only applies to responses, ignoring the involvement of the video.
\textbf{Right:} Our \textbf{P}rompt-aware \textbf{M}ulti-instance learning Video DPO (\method) overcomes these limitations \emph{by constructing a candidate rejected video set during training and automatically selects the appropriate rejected video based on the prompt to perform video-enhanced preference learning.}}
\label{fig:intro}
\vspace{-3.5mm}
\end{figure}

%% file: ICCV2025-Author-Kit-Feb/items/fig_intro_compare.tex
\begin{figure}[t]
\centering 
\includegraphics[width=0.48\textwidth,height=0.2\textheight]{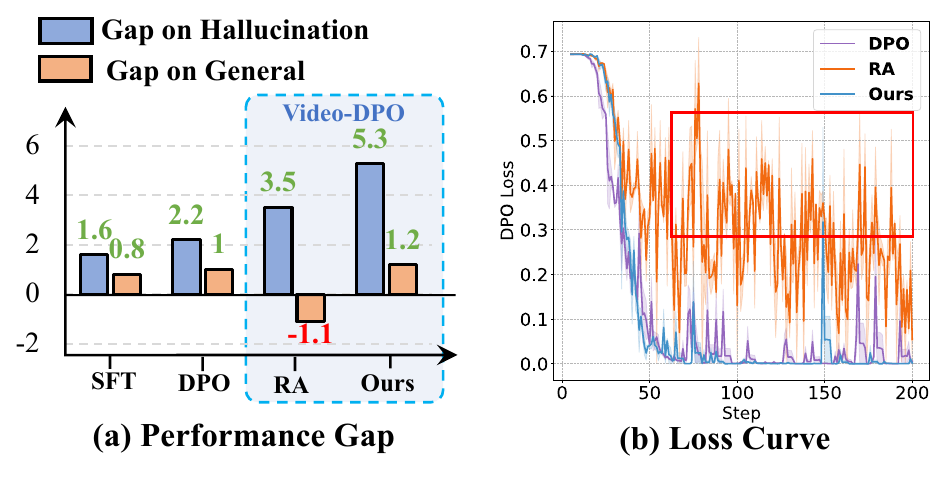}
\vspace{-1cm}
\caption{\textbf{Impact of Augmentations.} \textbf{RA:} Random video augmentation.  
\textbf{(a) Performance gap} with the base model (LLaVA-OV-7B~\cite{li2024llava}) and \textbf{(b) DPO loss curve} show that RA leads to performance degradation on general benchmarks and unstable training (\textcolor{red}{red box}). Performance gains and degradation are highlighted in \textcolor{mygreen}{green} and \textcolor{red}{-red}, respectively.  
In contrast, our \textbf{P}rompt-aware \textbf{M}ulti-instance learning Video DPO (\method) achieves superior performance on both hallucination and general benchmarks while ensuring stable training.}

\label{fig:intro_compare}
\vspace{-5.5mm}
\end{figure}

%% file: ICCV2025-Author-Kit-Feb/sec/2_related_work.tex
\section{Related Work}
\label{sec:related_work}

\vspace{-1mm}
\subsection{Hallucination in image and video MLLMs}\label{sec:related_halluciantion}

\vspace{-1mm}
\paragraph{Image Hallucination.} Multimodal large language models (MLLMs) often generate responses misaligned with visual content~\cite{liu2024survey}. Existing mitigation methods fall into two categories:
(i) Post-processing, which refines model outputs after generation~\cite{yin2024woodpecker, lee2024volcano, zhou2023analyzing, leng2024mitigating, zhao2024mitigating}, but adds inference overhead.
(ii) Data-driven preference alignment, where methods like Hallucidoctor~\cite{yu2024hallucidoctor} and LRV-Instruction~\cite{liu2023mitigating} improve dataset quality, while DPO-based approaches~\cite{li2023silkie, yu2024rlaifv, wang2024mdpo, rafailov2024direct} enhance alignment with visual content.

\paragraph{Video Hallucination.} Most video hallucination mitigation methods extend image-based strategies.
VISTA-LLAMA~\cite{ma2024vista} modifies attention mechanisms, improving alignment but limiting scalability for long videos. LLAVA-HOUND-DPO~\cite{zhang2024direct} applies DPO with GPT-4V-generated preference data, but relies on expensive, closed-source APIs~\cite{gpt4o}.
However, its reliance on offline preference data limits adaptability, would be outdated as the model evolves, and overlooks true video-response misalignment.

Unlike existing approaches, our method mitigates video hallucinations without modifying the architecture or relying on costly preference data generation, and conducts preference learning in an online manner.

\input{ICCV2025-Author-Kit-Feb/items/fig_motivation}

\vspace{-1mm}
\subsection{Direct preference optimization in MLLMs}
\vspace{-1mm}

Preference optimization~\cite{schulman2017proximal}, particularly Direct Preference Optimization (DPO)~\cite{rafailov2024direct,meng2024simpo}, has been widely used to align multimodal LLMs (MLLMs) with visual content~\cite{li2023silkie, yu2024rlaifv, wang2024mdpo, xie2024v, pi2024strengthening, zhu2024self, ouali2024clip, wang2024mdpo}. DPO relies on high-quality preference data, typically consisting of an image, a prompt, and two responses: one preferred (aligned with the image) and one rejected (more hallucinated). These annotations are often generated using multiple model responses~\cite{li2023silkie} or augmented from different seeds~\cite{yu2024rlaifv, zhang2024direct}, with labels from GPT-4o~\cite{gpt4o} or human.
To further improve alignment, some works ~\cite{xie2024v, fu2025chip, zhu2024self} modify the standard DPO objective by keeping responses fixed while perturbing the image to generate visually misaligned samples. This method has shown effectiveness in mitigating hallucinations in image-based MLLMs, as it enforces models to distinguish subtle visual differences.

Due to the additional temporal dimension, selecting augmentations for video DPO is non-trivial, generally suffers from false-rejected issues (see more in Section.~\ref{sec:empirical}).
To this end we introduce \method, a novel prompt multi-instance learning framework, significantly reducing false rejections while maintaining computational efficiency.

%% file: ICCV2025-Author-Kit-Feb/items/fig_motivation.tex
\begin{figure*}[t]
\centering 
\includegraphics[width=1.0\textwidth]{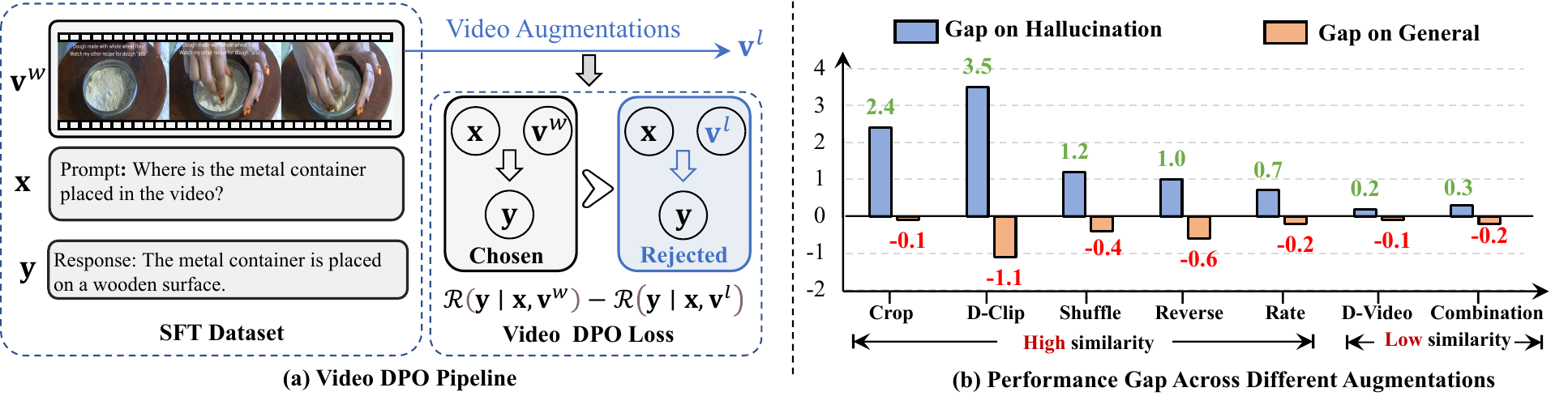}
\vspace{-1cm}
\caption{\textbf{(a) Video DPO (VDPO) Pipeline.} VDPO applies augmentations to the original clip $\mathbf{V}^w$ to generate the rejected clip $\mathbf{V}^l$, then optimized by the VDPO objective (Eq.~\ref{e:vdpo}).  
\textbf{(b) Performance Gap Across Augmentations.} The x-axis represents different augmentation strategies for generating rejected clips, while the y-axis shows the performance gap relative to the baseline (LLaVA-OV-7B~\cite{li2024llava}) across hallucination and general benchmarks (see Section~\ref{sec:setting}). Gains and degradations are highlighted in \textcolor{mygreen}{green} and \textcolor{red}{-red}, respectively.  
\textbf{High-similarity} augmentations retain semantic closeness to the original clip, whereas \textbf{low-similarity} ones introduce significant differences (see more in Section~\ref{sec:setting}).  
\emph{Different augmentation strategies have a substantial impact on VDPO performance, analyzed further in Section~\ref{sec:findings}.}}
\label{fig:motivation}
\vspace{-3.5mm}
\end{figure*}

%% file: ICCV2025-Author-Kit-Feb/sec/3_preliminary.tex
\section{Preliminary}\label{sec:preliminary}
In this section, we briefly review how existing approaches leverage direct preference optimization (DPO) to improve the performance of multimodal large language models (MLLMs), especially for mitigating hallucinations. 

\paragraph{Supervised Fine-tuning (SFT).} In this stage, we have the training SFT data $\mathcal{D}_{\mathrm{SFT}}=\{(\mathbf{x}, \mathbf{v}, \mathbf{y})\}$ where $x$, $v$ and $y$ are the instruction prompt, visual input and response respectively, Then, the MLLM with parameters $\pi_\theta$ is trained to maximize the log-likelihood of $y$ given $x$ and $v$ with the cross-entropy loss in the autoregressive manner~\cite{liu2023visual}. We follow the previous works~\cite{li2023silkie,yu2024rlaifv} that regard the model after the SFT stage as the reference model $\pi_\mathrm{ref}$.

\paragraph{Direct Preference Optimization (DPO).} Recent approaches leverage DPO to further enhance the response quality of MLLMs and reduce hallucinations. Specifically, the pair-wise preference data is formulated as $\mathcal{D}_{\mathrm{DPO}}=\{(\mathbf{x}, \mathbf{v}, \mathbf{y}^w, \mathbf{y}^w)\}$, where $\mathbf{y}^w$ is the chosen response which is preferred the rejected response $\mathbf{y}^l$.
Based on the constructed preference data, the DPO loss~\cite{yu2024rlaifv} can be formulated as:
\begin{equation}
 \mathcal{L}_{\text{dpo}}=- \log \sigma \left[ \beta* \left( \mathcal{R}\left(  \mathbf{y}^w \mid \mathbf{x}, \mathbf{v} \right) - \mathcal{R}\left(  \mathbf{y}^l \mid \mathbf{x}, \mathbf{v} \right) \right)\right],
\end{equation}
where $\beta$ controls the deviation $\pi_{\mathrm{\theta}}$ from $\pi_{\mathrm{ref}}$ during the optimization. $\mathcal{R}\left(  \mathbf{y}^w \mid \mathbf{x}, \mathbf{v} \right)$ and $\mathcal{R}\left(  \mathbf{y}^l \mid \mathbf{x}, \mathbf{v} \right)$ are the chosen and rejected rewards, defined as follows:
\begin{equation}
    \mathcal{R}\left(  \mathbf{y}^w \mid \mathbf{x}, \mathbf{v} \right) = \log \left(\frac{\pi_{\mathrm{\theta}}\left(\mathbf{y}^w \mid \mathbf{x}, \mathbf{v}\right)}{\pi_{\mathrm{ref}}\left(\mathbf{y}^w \mid \mathbf{x}, \mathbf{v}\right)}\right),
\end{equation}
where $\pi_{\mathrm{ref}}$ is the fixed reference model, $\pi_{\mathrm{\theta}}$ is the trained model during the DPO.

%% file: ICCV2025-Author-Kit-Feb/sec/4_vdpo.tex
\section{Video Preference Learning}\label{sec:empirical}

\vspace{-0.5mm}
\subsection{Overall Framework}\label{sec:vdpo}
\vspace{-1mm}

Direct Preference Optimization (DPO) has been applied to mitigate hallucinations in VLLMs by preconstructing preference data (correct vs. hallucinated responses). However, it suffers from high annotation costs, offline data limitations, and response-focused learning biases. To overcome these issues, we introduce \textbf{Video DPO (VDPO)}, which removes the need for manually curated preference data by enforcing preference learning directly on videos.

As shown in Fig.~\ref{fig:motivation}~(a), VDPO consists of three steps:  
(i) Sampling training examples $\{\mathbf{x}, \mathbf{y}, \mathbf{v}^w\}$ from SFT datasets;  
(ii) Applying video augmentations to the original video clip $\mathbf{v}^w$ to generate the rejected video clip $\mathbf{v}^l$;  
(iii) Optimizing the model (e.g., LLaVA-OV-7B~\cite{li2024llava}) using the VDPO objective:  

\vspace{-4mm}
\begin{equation}
 \mathcal{L}_{\text{vdpo}}=- \log \sigma \left[ \beta* \left( \mathcal{R}\left(  \mathbf{y} \mid \mathbf{x}, \mathbf{v}^w \right) - \mathcal{R}\left(  \mathbf{y} \mid \mathbf{x}, \mathbf{v}^l \right) \right)\right].
 \label{e:vdpo}
\end{equation}

\subsection{Augmentation: A Bottleneck in VDPO}\label{sec:bottleneck}

As Eq.~\ref{e:vdpo} shows, the only variable in VDPO is the video input, meaning that the construction of the rejected clip plays a crucial role in its effectiveness. A key question arises: \emph{What kinds of video augmentations are optimal for VDPO?}

To answer this, we conduct an empirical study analyzing the impact of different augmentations.

\subsubsection{Experiment Setup}\label{sec:setting}

\paragraph{Training Data.} We sample 10K SFT examples from \textit{Temporal-Bench}~\cite{cai2024temporalbench} and \textit{LLaVA-Video-178k}~\cite{zhang2024videoinstructiontuningsynthetic} to evaluate performance across temporal and visual domains.

\paragraph{Augmentation Strategies.} We apply the following video augmentations to construct rejected clips: \textbf{(1) Crop} -- randomly cropping less than 20\% of the frame from the original video; \textbf{(2) D-video} -- randomly selecting clips from another video from the training data; \textbf{(3) D-Clip} -- selecting another clip from the same long video with the original video clip. \textbf{(4) Shuffle} -- shuffling the temporal order of the original video clip; \textbf{(5) Reverse} -- reversing the original temporal order; \textbf{(6) Rate} -- sampling frames from the original video clip at a different frame rate;  \textbf{(7) Combination} -- applying two ways sampled from previous augmentation strategies.

For further analysis, we categorize the above augmentations into two groups: \emph{\textbf{(i) High-similarity Augmentations:} augmentations that do not significantly alter the semantic content of the original video clip, resulting in a high similarity between the augmented and original clips.} \emph{\textbf{(ii) Low-similarity Augmentations:} augmentations that significantly alter the original semantic content, making it difficult or impossible to recognize the original scene or objects.} 

\paragraph{Evaluation Metrics.} We evaluate models on:  
- {General Benchmarks}: Average performance on \textit{TempCompass}~\cite{liu2024tempcompass}, NextQA~\cite{xiao2021next} and \textit{Video-MME}~\cite{liu2024tempcompass}.  
- {Hallucination Benchmarks}: Five sub-tasks from \textit{VideoHallucer}~\cite{wang2024videohallucer}.  

\input{ICCV2025-Author-Kit-Feb/items/fig_analysis}

\subsubsection{Experimental Findings}\label{sec:findings}

\paragraph{High-Similarity Augmentations Improve Hallucination Mitigation.}  
As shown in Fig.~\ref{fig:motivation}~(b), VDPO models trained with high-similarity augmentations outperform those using low-similarity augmentations. This is because high-similarity augmentations generate subtle yet distinguishable negative samples, forcing the model to focus on fine-grained video-response misalignment. For example, {D-Clip achieves a +3.5\% improvement} in hallucination benchmarks.

\paragraph{Low-Similarity Augmentations Provide Minimal Benefit.}  
Low-similarity augmentations create rejected clips with obvious semantic gaps, making it trivial for the model to distinguish between the original and rejected samples. As a result, these augmentations fail to provide meaningful supervision. Fig.~\ref{fig:analysis}~(a) shows that VDPO loss converges too quickly (\textcolor{orange}{orange curve}), indicating ineffective learning.

\paragraph{Instability of High-Similarity Augmentations.}  
Despite their benefits, high-similarity augmentations introduce training instability:

\begin{itemize}
    \item \textbf{Performance sensitivity} – Gains vary widely by augmentation (e.g., {D-Clip: +3.5\%}, {Reverse: +1.0\%}).
    \item \textbf{General benchmark degradation} – High-similarity augmentations often cause a performance drop on general tasks (e.g., {D-Clip: -1.1\%}).
    \item \textbf{Slow and unstable loss convergence} – As shown in Fig.~\ref{fig:analysis}~(a), high-similarity augmentations (\textcolor[HTML]{4292c9}{blue curve}) exhibit large variance and slower convergence.
\end{itemize}

\input{ICCV2025-Author-Kit-Feb/items/fig_clean_mixed}

\paragraph{False-Rejected Issue in VDPO.}  
The effectiveness of an augmentation varies depending on the prompt. As shown in Fig.~\ref{fig:analysis}~(b), applying {Reverse} to generate a rejected clip is effective for {Prompt-A} (where event order matters) but fails for {Prompt-B} (where object positions remain unchanged). Such false-rejected samples disrupt VLLM alignment, leading to instability.

To validate this,  we design a controlled experiment by curating temporal-related questions from the original SFT dataset. Specifically, we leverage the GPT-4o API~\cite{gpt4o} to identify questions involving temporal reasoning (e.g., direction prediction, velocity estimation, or sequential understanding), resulting in 3K samples termed Clean data. For fair comparison, we randomly select 3K samples from the original dataset as Mixed data.
Subsequently, we apply Shuffle augmentation to train two VDPO models: one on the Clean data (resulting in the Shuffle-Clean model) and the other on the Mixed data (resulting in the Shuffle-Mixed model).
In Fig.~\ref{fig:clean_mixed}, we report the performance gap of these models compared to a baseline model on hallucination and general benchmarks, as well as examining the training loss.
Our results demonstrate that Shuffle-Clean significantly mitigates the instability of VDPO. This improvement stems from the fact that, for temporal-related questions in the Clean data, shuffled video clips are likely true-rejected, as shuffling disrupts the temporal information essential for accurate responses.
Hence, tailoring augmentation strategies to the questions to avoid false-rejected is critical for addressing instability in VDPO training.

%% file: ICCV2025-Author-Kit-Feb/items/fig_analysis.tex
\begin{figure}[t]
\centering 
\includegraphics[width=0.48\textwidth,height=0.21\textheight]{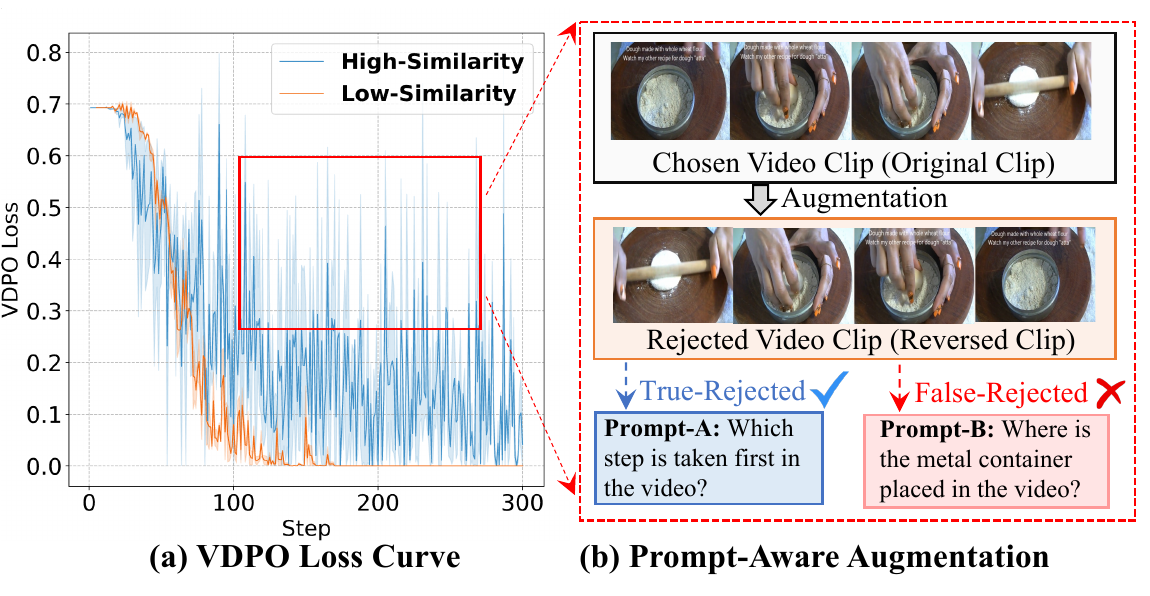}
\vspace{-0.8cm}
\caption{\textbf{(a) VDPO loss curve.} The solid line depicts the average loss, with the shaded area showing the variance. \emph{VDPO loss trained with high-similarity augmentations exhibits greater variance, whereas low-similarity augmentations enable smoother convergence.}
\textbf{(b) Prompt-aware augmentation.} The impact of augmentations varies by prompts,~\ie, \emph{the augmented rejected video clip may have the same response as the chosen clip for some prompts}. We term these generated clips as false-rejected.
}
\label{fig:analysis}
\vspace{-3.5mm}
\end{figure}

%% file: ICCV2025-Author-Kit-Feb/items/fig_clean_mixed.tex
\begin{figure}[t]
\centering 
\includegraphics[width=0.48\textwidth,height=0.16\textheight]{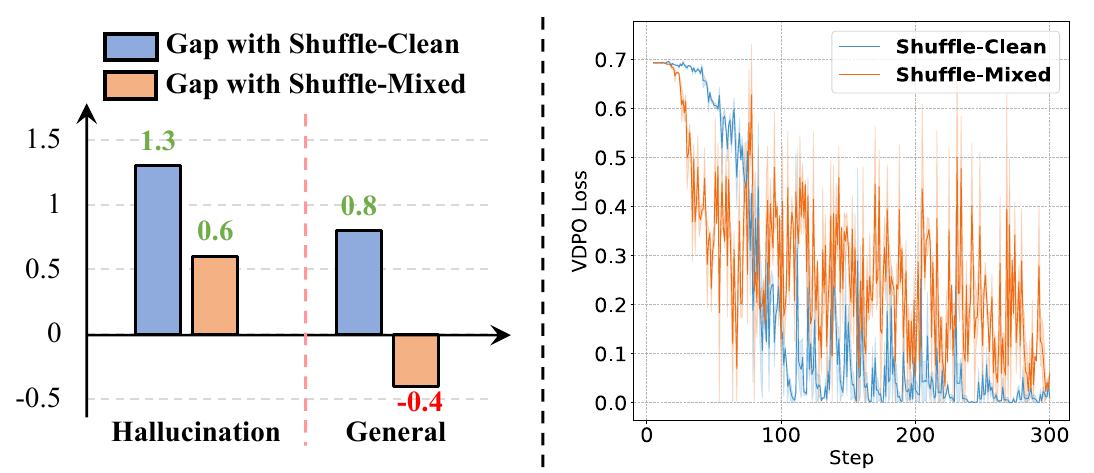}
\vspace{-0.8cm}
\caption{\textbf{Prompt-aware augmentation stabilizes VDPO.} Comparative analysis on both the performance gap (\textbf{Left}) and the loss curve (\textbf{Right}) reveals \emph{selecting appropriate augmentations based on the questions can help mitigate VDPO instability.} Specifically, \textcolor[HTML]{4292c9}{Shuffle-Clean} and \textcolor{orange}{Shuffle-Mixed} represent VDPO training with the Shuffle augmentation on Clean and Mixed data, respectively (see Section~\ref{sec:findings} for details). \textcolor[HTML]{4292c9}{Shuffle-Clean} consistently {reduces hallucination} and maintains generalizability, whereas \textcolor{orange}{Shuffle-Mixed} degrades both the general tasks and optimization stability.
}
\label{fig:clean_mixed}
\vspace{-3.5mm}
\end{figure}


%% file: ICCV2025-Author-Kit-Feb/sec/5_method.tex
\input{ICCV2025-Author-Kit-Feb/items/fig_milvdpo}

\section{\method}\label{sec:method}

\paragraph{The VDPO dilemma: hallucination vs. stability.}
While VDPO mitigates hallucinations by enforcing preference learning on video inputs, it faces a fundamental trade-off in constructing rejected clips. On one hand, rejected clips should maintain high semantic similarity with the original to help the model detect subtle inconsistencies (Fig.~\ref{fig:motivation}(b)). On the other hand, overly similar rejected clips can lead to false-rejected samples, introducing alignment noise and destabilizing training (Fig.~\ref{fig:analysis}). 

Our experiments in Fig.~\ref{fig:clean_mixed} suggest that prompt-dependent high-similarity augmentations can reduce false-rejected samples, but manually selecting these augmentations via human curation or LLM-based filtering is costly.

\paragraph{Prompt-aware Multi-instance learning formulation.}
To address this dilemma without relying on human/LLM pre-selection, we propose constructing a candidate set $\mathcal{V} = \{ \mathbf{v}_i^l \}_{i=1}^N$ of $N$ augmented video clips for each video-prompt pair $\{\mathbf{x}, \mathbf{v}^w\}$. Instead of relying on a single rejected clip, multiple candidates substantially increase the chance that \textbf{at least one true-rejected video clip} in $\mathcal{V}$ aligns with the given prompt. As shown in Fig.~\ref{fig:milvdpo} (a), among all the augmented clips $\mathcal{V}$, the one augmented by D-Clip $\mathbf{v}_2^l$ is indeed the true-rejected clip matched with the prompt $\mathbf{x}$.

Notably, this candidate set aligns structurally with Multiple Instance Learning (MIL)~\cite{zhou2004multi}, where $\mathcal{V}$ serves as a bag containing both positive (true-rejected) and negative (false-rejected) instances. Consequently, we reformulate the VDPO objective (Eq.~\ref{e:vdpo}) into a Prompt-aware MIL VDPO (\method) as follows:
\begin{equation}
\begin{aligned}
     & \mathcal{L}_{\text{pami-vdpo}}= \\
     & - \log \sigma \left[ \beta * \left( \mathcal{R}\left(  \mathbf{y} \mid \mathbf{x}, \mathbf{v}^w \right) - \textcolor{myblue}{\sum_{\mathbf{v}_i^l \in \mathcal{V}} \alpha_i *}\mathcal{R}\left(  \mathbf{y} \mid \mathbf{x}, \textcolor{myblue}{\mathbf{v}_i^l} \right) \right)\right],
     \label{e:qmvdpo}
\end{aligned}
\end{equation}
where $\alpha_i \in [0,1]$ is a prompt-aware confidence score for clip $\mathbf{v}_i^l$. A higher $\alpha_i$ suggests a stronger likelihood that $\mathbf{v}_i^l$ is a pseudo true-rejected clip, while a lower $\alpha_i$ implies potential noise (false-rejected). This MIL-based weighting scheme selectively optimizes high-confidence clips and suppresses noisy ones.

\input{ICCV2025-Author-Kit-Feb/items/tab_halluciation_v2}

\paragraph{Tailored designs for effectiveness.}
To ensure that our MIL formulation effectively selects useful rejected samples, we introduce a \textbf{close-to-far} strategy, which balances the trade-off between retaining semantic similarity and avoiding false-rejected clips.

\textit{Close construction:} We use diverse high-similarity augmentations to generate the rejected set $\mathcal{V}$. Because all clips in $\mathcal{V}$ remain semantically close to the original, they help mitigate hallucinations (refer to Section~\ref{sec:findings} and Fig.~\ref{fig:motivation}(b)); meanwhile, diversity among these clips increases the likelihood of discovering true-rejected instances.

\textit{Far selection:} We select the augmented clip that differs the most from the original clip’s LLM output as the true-rejected clip. Since LLM outputs depend on both the prompt and the video clip, a large output divergence indicates that the augmentation meaningfully affects the response. By emphasizing the farthest clip and down-weighting others, we reduce the risk of false-rejected noise.

Specifically, we measure the distance between outputs using the Jensen-Shannon (JS) divergence. Let $\mathbf{p}^w$ be the MLLMs’ output logits for input $\{\mathbf{x}, \mathbf{v}^w\}$ and $\mathbf{p}^l_i$ be the logits for $\{\mathbf{x}, \mathbf{v}^l_i\}$. After converting logits to probabilities via softmax, we compute:
\begin{equation}
    d_i = \text{JS}\bigl(\text{softmax}(\mathbf{p}^w) \,\|\, \text{softmax}(\mathbf{p}^l_i)\bigr).
    \label{e:distance}
\end{equation}

We then derive prompt-aware confidence weights through a softmax over $d_i$ values:
\begin{equation}
    a_i=\frac{\exp \left(d_i\right)}{\sum_{j=1}^N \exp \left(d_j\right)},\
    \label{e:alpha}
\end{equation}

Notably, our framework introduces zero additional parameters and requires no architectural changes, making it straightforward to integrate into existing LLM training pipelines in a plug-and-play fashion.

%% file: ICCV2025-Author-Kit-Feb/items/fig_milvdpo.tex
\begin{figure}[t]
\centering 
\includegraphics[width=0.48\textwidth,height=0.28\textheight]{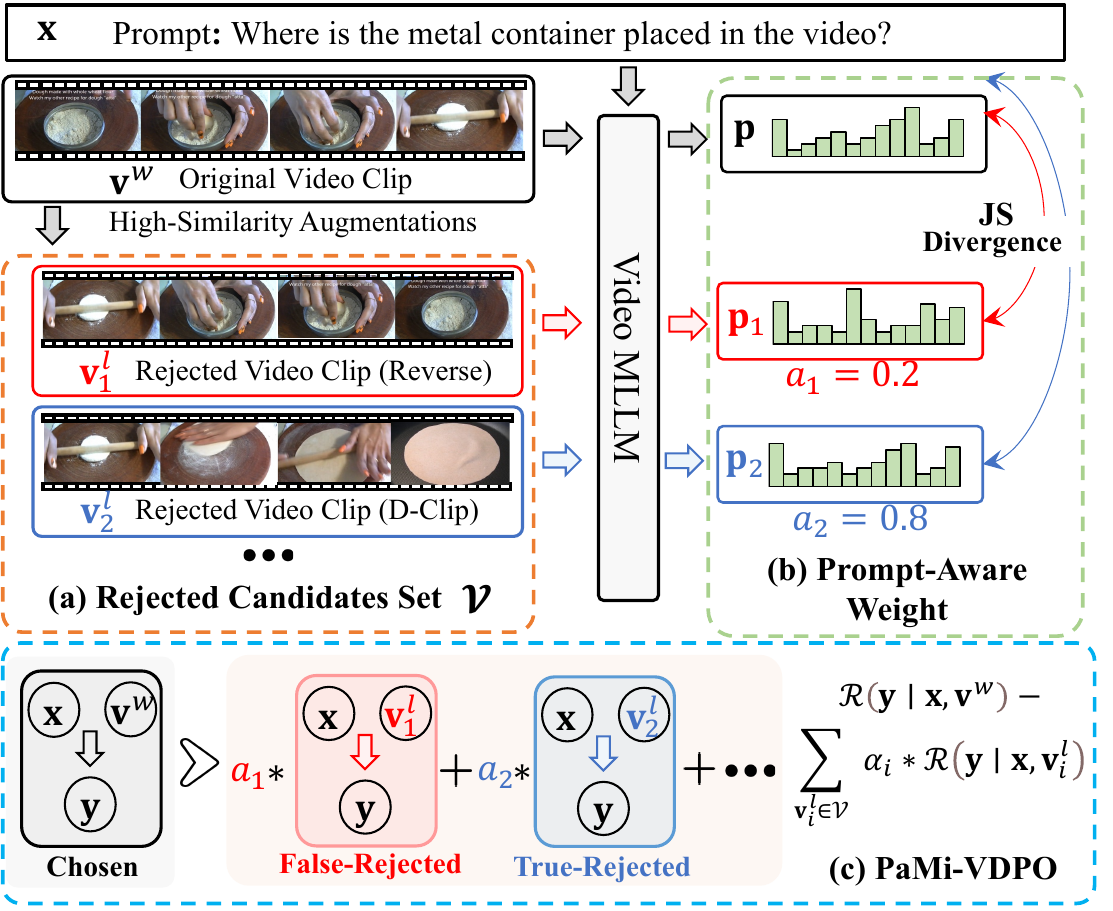}
\vspace{-0.8cm}
\caption{\textbf{Framework of our method.} Our method addresses VDPO's instability caused by false-rejected clips through three key innovations: \textbf{(a) Rejected candidate set construction $\mathcal{V}$} that guarantees at least one is the true-rejected clip via multiple video augmentations;
\textbf{(b) Prompt-Aware Weighting} that computes the Jensen-Shannon divergence between the LLM outputs of the original clip and those in $\mathcal{V}$, enabling us to identify the high-confidence rejected clip during training, without adding extra parameters;
\textbf{(c) \method~Objective} emphasizes the high-confidence rejected candidates (\textcolor[HTML]{4292c9}{blue boxes}) while suppressing other samples (\textcolor{red}{red boxes}) during the preference learning.
}
\label{fig:milvdpo}
\vspace{-3.5mm}
\end{figure}


%% file: ICCV2025-Author-Kit-Feb/items/tab_halluciation_v2.tex
\begin{table}[t]
\center
\setlength{\tabcolsep}{5pt}
\Large
 \begin{adjustbox}{max width=1.0\linewidth}
\begin{tabular}{l|cclccc}
\Xhline{1.5pt}
 &  \multicolumn{3}{c}{{\textbf{VideoHallucer}}} & \multicolumn{3}{c}{\textbf{EventHallusion}} \\
\cmidrule(lr){2-4} \cmidrule(lr){5-7}
\textbf{Models} &  Basic & Hallu & Overall &Binary & Desc & Avg \\ 
\midrule
VideoChatGPT~\cite{maaz2023video} & 92.8& 10.4 & 6.4 & 47.19 & 11.6 & 29.4\\
VideoChat2~\cite{li2024mvbench} &  29.7 & 25.8 & 7.8 & 37.90 &7.8 & 22.9\\
LLaMA-VID~\cite{li2024llama} &  89.9 & 26.6 & 21.0&  66.01 & 18.6 & 42.3 \\
Video-LLaVA~\cite{lin2023video} & 95.1 & 20.3 & 17.8 &  50.61 & 16.9 &33.8 \\
\midrule
LLaVA-OV~\cite{li2024llava} & 84.1 & 60.8 & 48.5 & 61.3 & 33.2 & 47.3\\
+ SFT & \textbf{85.1} &  61.1 & 50.1 {\small \textbf{\textcolor{tabfirst}{(+1.6)}}} & 66.0 & 30.7 & 48.4 {\small \textbf{\textcolor{tabfirst}{(+1.1)}}}\\
+ DPO & 81.9 & 66.2 & 51.2 {\small \textbf{\textcolor{tabfirst}{(+2.7)}}} & \textbf{69.9} & 31.7 & 50.8 {\small \textbf{\textcolor{tabfirst}{(+3.5)}}}\\
  \rowcolor{tabfirstred!20}
+ VDPO & 75.5 & 67.9 & 51.1 {\small \textbf{\textcolor{tabfirst}{(+2.6)}}} & 67.2 & 32.7 & 50.0 {\small \textbf{\textcolor{tabfirst}{(+2.7)}}}\\
 \rowcolor{lightblue!20}
+ \method (ours) & {82.8} & \textbf{69.9} & \textbf{53.8} {\small \textbf{\textcolor{tabfirst}{(+5.3)}}} & \underline{67.9} & \textbf{38.5} & \textbf{53.2} {\small \textbf{\textcolor{tabfirst}{(+5.9)}}}\\
\midrule
  \multicolumn{7}{l}{\textit{Proprietary models}} \\
    \rowcolor{mygray}
Gemini-1.5-Pro~\cite{team2024gemini} & 83.6 & 42.3 & 37.8 & 80.2 & 49.6& 64.9 \\
    \rowcolor{mygray}
GPT-4o~\cite{gpt4o}  & 75.1 & 74.2 & 53.5& 84.1 & 56.2 & 70.2 \\
\bottomrule
\end{tabular}
\end{adjustbox}
\vspace{-0.5em}
\caption{\textbf{Comparison with the stat-of-the-art MLLMs on hallucination benchmarks.} The LLM size of all models is around 7B. For all results, the higher value indicates a better performance. `AVG' is the average performance. The best and the second-best values are marked in {\bf Bold} and \underline{Underline}. The performance gains is highlighted in \textcolor{tabfirst}{(+blue)}. \emph{Trained with the same data, our \method~outperforms SFT, DPO and VDPO across different backbones and benchmarks.}}
\label{tab:hallucination}
\vspace{-5mm}
\end{table}

%% file: ICCV2025-Author-Kit-Feb/sec/6_experiments.tex
\input{ICCV2025-Author-Kit-Feb/items/tab_general}
\section{Experiments}

\subsection{Experimental settings}

\paragraph{Implementation Details.} We use LLaVA-OV-7B~\cite{li2024llava} as the backbone and train for 3 epochs with a learning rate of $5e-7$ and batch size of 1. The number of sampled frames is 8 during training and 32 for inference. To improve efficiency, we apply $2\times2$ average-pooling to the visual features of each frame. The candidate set size is fixed at 2 for an optimal balance between efficiency and performance. We use D-Clip (or Crop if D-Clip is unavailable) and Shuffle augmentations to construct the set.
For further analysis, refer to Section~\ref{sec:ablation} and \textbf{\sm}.

\paragraph{Training Data.} We sample around $10K$ SFT data examples from Temporal-bench~\cite{cai2024temporalbench} and LLaVA-Video-178k\cite{zhang2024videoinstructiontuningsynthetic}, the same as Section~\ref{sec:setting}. 

\paragraph{Evaluation Benchmarks.} We evaluate our method on two hallucination (VideoHallucer~\cite{wang2024videohallucer}, EventHallusion~\cite{zhang2024eventhallusion}) and three general (NextQA~\cite{xiao2021next},Tempcompass (Tempcom)~\cite{liu2024tempcompass} and Video-MME (VMME)~\cite{fu2024video}) benchmarks. For hallucination benchmarks, we use their official evaluation code, while LMMS-Eval~\cite{li2024lmms} for evaluating general benchmarks.

\subsection{Comparison with state-of-the-art methods}

\input{ICCV2025-Author-Kit-Feb/items/tab_set}

We apply our method to a powerful VLLM,~\ie, LLaVA-OV~\cite{li2024llava}.
For fair comparison, we also train the backbone using the same data by the supervised fine-tuning (SFT) and the standard DPO.
The standard DPO requires the chosen and rejected response pairs for each data sample.
To this end, for each sample, we first generate multiple responses by inputting different augmented videos, which are then annotated by a powerful model,~\ie, GPT-4o~\cite{gpt4o}.
Refer to the \textbf{supplementary material} for more details.
\emph{Note, due to the lack of publicly available results where all models are evaluated on both hallucination and general benchmarks, we report them separately while ensuring our baseline (LLaVA-OV) is included in both for reference.}

\paragraph{Hallucination benchmarks.} Table~\ref{tab:hallucination} presents the hallucination performance comparison of our methods with other training methods,~\ie SFT, DPO, and VDPO.
Note that we keep the training data to be the same for SFT, DPO, VDPO and our \method.
We can observe two main findings: \emph{(i) preference learning outperforms the standard SFT on both hallucination benchmarks;} 
 \emph{(ii) Our \method~shows better performance than both DPO and native VDPO across different backbones.}
More importantly, unlike DPO, our method does not require pre-constructing preference data, reducing usage costs while also avoiding hallucinations introduced by data construction.

\paragraph{General benchmarks.} Table~\ref{tab:general} compares ours with different learning methods on the general benchmarks.
Although SFT shows limited gains on the hallucination benchmark, it consistently improves general tasks. In contrast, DPO and VDPO suffer notable drops, especially VDPO, which declines by 2\% on Video-MME due to false-rejected samples disrupting alignment. By avoiding this issue, our method achieves stable improvements on general benchmarks.

\subsection{Ablation Study}\label{sec:ablation}

In this section, we conduct the extensive experiments to ablate the effect of our proposed modules.
We use the LLaVA-OV-7B~\cite{li2024llava} as our baseline model.
As analyzed in Section~\ref{sec:method}, the effectiveness of \method~relies on the two key components: the constructed rejected candidates set $\mathcal{V}$ and prompt-dependent weight $\alpha_i$.
We ablate the two components in Table~\ref{tab:ablations} to prove that the effectiveness of the close-to-far strategy.

\subsubsection{Rejected Candidates Set}
\label{sec:ablation_set}

As analyzed in Section~\ref{sec:method}, the augmentations used to generate $\mathcal{V}$ have two key properties: high similarity and diversity, which are ablated in Table~\ref{tab:augtype} and Table~\ref{tab:diversity}, respectively.

\paragraph{Augmentation Group.}  
Table~\ref{tab:augtype} compares three augmentation settings for constructing $\mathcal{V}$ (details in Section~\ref{sec:setting}):  
(i) High-similarity: only high-similarity augmentations;  
(ii) Low-similarity: only low-similarity augmentations;  
(iii) Combination: a mix of both, with equal proportions.  
For all settings, we fix $|\mathcal{V}| = 2$ (further ablations in the \textbf{\sm}~confirm this suffices for strong performance).  

Results show that high-similarity augmentations achieve the best performance, aligning with findings in Section~\ref{sec:findings} and Fig.~\ref{fig:motivation}~(b).  
In the combination setting, the farthest clip (typically low-similarity) is more likely to be selected due to prompt-aware weighting.  
However, our preference learning across the entire candidate set mitigates the noise from low-similarity clips, leading to a significantly higher performance gain than using only low-similarity augmentations (\eg, +2.4\% vs. +0.2\% on the hallucination benchmark), demonstrating the robustness of our method.

\paragraph{Augmentation Diversity.}  
In Table~\ref{tab:diversity}, we evaluate the effect of the augmentation diversity: (i) Temporal: only augment clips in the temporal dimension (Shuffle, Reverse, Rate); (ii) only augment clips in the visual dimension (Crop, D-Clip); (iii) Mixed: some clips are augmented in the temporal dimension while others are in the visual dimension.  
All augmentations in this study are high-similarity.

Results indicate that the Mixed setting achieves the best performance, as leveraging both temporal and visual augmentations increases the likelihood of generating true-rejected samples that align with different prompt types.  

\subsubsection{Prompt-aware Weight}
\label{sec:ablation_weight}

The formulation in Eq.~\ref{e:alpha} shows that the value of \weight~$\alpha_i$ for clip $\mathbf{v}_i^l$ is related to the distance $d_i$ of $\mathbf{v}_i^l$ to the original clip $\mathbf{v}_w$, which is ablated in the following.

\input{ICCV2025-Author-Kit-Feb/items/tab_generalization}

\paragraph{Weight $\alpha_i$ in Eq.~\ref{e:alpha}.}  
Table~\ref{tab:weigth} compares three different strategies for computing $\alpha_i$:  
(i) {Far}: $\alpha_i$ increases as $d_i$ increases (Eq.~\ref{e:alpha});  
(ii) {Near}: $\alpha_i$ decreases as $d_i$ increases (by replacing $d_i$ with $-d_i$ in Eq.~\ref{e:alpha});  
(iii) {Equal}: $\alpha_i = 1/N$, assigning equal weight to all rejected clips.  

The results show that the Far strategy achieves the best trade-off between hallucination mitigation and general tasks. In contrast, the {Near} strategy performs worse in both cases, as it selects rejected clips that are too similar to the original, making preference learning less effective in distinguishing meaningful differences and degrading alignment.  

Although the {Equal} strategy does not perform as well as {Far}, it still achieves a notable improvement in hallucination tasks while maintaining general task performance. This suggests that leveraging a diverse set of rejected clips acts as a form of soft regularization, enhancing training stability.  

\paragraph{Distance $d_i$ in Eq.~\ref{e:distance}.} In Table~\ref{tab:distance}, we compare the effect of different ways to compute the distance relative to the original clip.
Specifically, the Visual way means computing the cosine similarity between the visual features of the original clip $\mathbf{v}^w_i$ and the current rejected clip $\mathbf{v}^l_i$. For this, we apply the average-pooling to the output frame features for each video clip from the vision encoders of VLLMs.
The output way is our default setting defined in Eq.~\ref{e:distance}, which considers the information from both prompts and the video clips.
Results show that the output way outperforms the visual way clearly on the hallucinations benchmarks,~\eg, $5.3\%$ vs. $3.7\%$.
For general tasks, both ways achieve stable performance.

\subsection{Generalization}
\vspace{-1mm}

We evaluate the generalization of our method on both dataset and backbone aspects in Table~\ref{tab:generalization}.

\paragraph{Dataset.}  
To assess generalization beyond our training set, we compare our method with DPO on an additional public video preference dataset, LLaVA-HOUND-17k~\cite{zhang2024direct}, which samples videos from WebVid~\cite{bain2021frozen}, VIDAL~\cite{zhu2023languagebind}, and ActivityNet~\cite{caba2015activitynet}, with preference data generated by GPT-4V.  

As shown in Table~\ref{tab:llava_hound}, both DPO and \method~achieve only marginal improvements on general benchmarks, likely because LLaVA-HOUND-17k overlaps with the SFT data used for LLaVA-OV. However, on both hallucination benchmarks, our method consistently outperforms DPO. Notably, despite LLaVA-HOUND-17k containing 17K preference-labeled samples—more than the 10K SFT samples used in Table~\ref{tab:hallucination}—the performance gains from training on LLaVA-HOUND-17k are smaller. This highlights a key advantage of \method: it enables direct preference learning on high-quality SFT data without requiring costly pre-constructed preference labels.

\paragraph{VLLMs.}  
Besides LLaVA-OV~\cite{li2024llava}, we further evaluate our method on another VLLM to test its robustness across architectures. Table~\ref{tab:general_model} compares our method with DPO on LongVA-7B~\cite{zhang2024long}. The results show that \method~consistently outperforms DPO across different models and benchmarks, demonstrating its adaptability to varying backbone architectures.

\input{ICCV2025-Author-Kit-Feb/items/fig_visual}

\subsection{Qualitative Results}  

To highlight the advantages of our method over standard DPO, we present a visual comparison in Fig.~\ref{fig:visual}, using two examples from VideoHallucer (\textbf{Top}) and EventHallusion(\textbf{Bottom}).  
In the top example, both methods receive the same prompt, but the main object in the video changes from a metal box to a fabric box. Our method correctly identifies this change, while DPO fails to adjust its response accordingly.  
In the bottom example, DPO generates a detailed yet misleading description, assuming that the two dogs are happily playing—an inference likely influenced by LLM biases. \emph{This suggests that standard DPO prioritizes linguistic patterns over actual video understanding.} In contrast, our method directly learns preference signals from video content, allowing it to focus on visual details and produce more accurate responses.

%% file: ICCV2025-Author-Kit-Feb/items/tab_general.tex
\begin{table}[t]
\center
\setlength{\tabcolsep}{5pt}
\Large
 \begin{adjustbox}{max width=1.0\linewidth}
\begin{tabular}{l|ccc|l}
\Xhline{1.5pt}
&   {\textbf{NextQA}} & {\textbf{{VideoMME}}} & {\textbf{Tempcom}} \\
 \textbf{Models} &  (mc) &  (wo sub.) & (mc) & \textbf{AVG} \\
\midrule
InternVL2-8B~\cite{chen2024far} & 65.3 & 54.0 & -&-\\
VILA-40B~\cite{lin2024vila} &  67.9 & 60.1 & - &-\\
LongVA-7B~\cite{zhang2024long}  & 68.3 & 52.6 & - \\
LLaVA-OV~\cite{li2024llava}  & 79.4 & 58.2 & 64.8 & 67.5\\
\midrule
LLaVA-OV$^*$ & 77.8 & 58.4 & 64.8 & 67.0 \\
+SFT & 78.8 & 58.9 & 66.2 & 68.0 {\small \textbf{\textcolor{tabfirst}{(+1.0)}}}\\
+DPO & 79.2 & 58.8 & 64.7 & 67.6 {\small \textbf{\textcolor{tabfirst}{(+0.6)}}} \\
  \rowcolor{tabfirstred!20}
+VDPO & 77.8 & 56.7 & 63.3 & 65.9 {\small \textbf{\textcolor{red}{(-1.1)}}} \\
 \rowcolor{lightblue!20}
+\method~(ours) & 79.7 & 58.5 & 66.4 & 68.2 {\small \textbf{\textcolor{tabfirst}{(+1.2)}}}\\
\bottomrule
\end{tabular}
\end{adjustbox}
\vspace{-0.5em}
\caption{\textbf{Comparison with the stat-of-the-art MLLMs on general benchmarks.} For all results, the higher value indicates a better performance. `$^*$' indicates the results reproduced by ourselves. The performance degradation is highlighted in \textcolor{red}{(-red)}. \emph{Compared with the standard DPO or VDPO, our \method~would not degrade the performance of general benchmarks, due to reducing the false-rejected clips.}}
\label{tab:general}
\vspace{-5mm}
\end{table}

%% file: ICCV2025-Author-Kit-Feb/items/tab_set.tex
\begin{table*}[t]
\vspace{-.2em}
\centering
\subfloat[
\textbf{Augmentation group}. $\mathcal{V}$ generated by only high-similarity augmentations ensures all clips are close-semantic with the original clip.
\label{tab:augtype}
]{
\begin{minipage}{0.24\linewidth}{\begin{center}
\tablestyle{3pt}{1.05}
\begin{tabular}{y{50}x{20}x{20}}
group & {\it \textbf{Halluc}} & {\it \textbf{General}} \\
\shline
High-similarity & \baseline{\textbf 5.3} & \baseline{\textbf 1.2} \\
Low-similarity &  0.2 & \textcolor{red}{-0.1} \\
Combination & 2.4 & 0.3 \\
\end{tabular}
\end{center}}\end{minipage}
}
\hspace{1em}
\subfloat[
\textbf{Augmentation diversity}. Diverse high-augmentations enlarge the chance of true-rejected samples related to the questions.
\label{tab:diversity}
]{
\centering
\begin{minipage}{0.24\linewidth}{\begin{center}
\tablestyle{3pt}{1.05}
\begin{tabular}{y{40}x{20}x{20}}
type & {\it \textbf{Halluc}} & {\it \textbf{General}} \\
\shline
Temporal & 3.5 & 1.1 \\
Visual & 3.9 &  0.7\\
Mixed & \baseline{\textbf 5.3} & \baseline{\textbf 1.2} \\
\end{tabular}
\end{center}}\end{minipage}
}
\hspace{1em}
\subfloat[
\textbf{Weight $\alpha_i$}. Enhance weight based on the farthest clip ($\alpha_i$ is proportional to $d_i$) is better for avoiding false-rejected issue. 
\label{tab:weigth}
]
{
\begin{minipage}{0.2\linewidth}{\begin{center}
\tablestyle{3pt}{1.05}
\begin{tabular}{y{45}x{20}x{20}}
$\alpha_i$ in Eq.~\ref{e:alpha} & {\it \textbf{Halluc}} & {\it \textbf{General}} \\
\shline
Far ($d_i$) & \baseline{\textbf{5.3}} & \baseline{\textbf 1.2}\\
Near ($-d_i$) & 3.8 & \textcolor{red}{-0.7}\\
Equal ($\frac{1}{N}$) & 4.6 & 0.4 \\
\end{tabular}
\end{center}}\end{minipage}
}
\centering
\hspace{1em}
\subfloat[
\textbf{$d_i$ computation}. Computing $d_i$ based on LLM's output $\mathbf{p}_i$ is better than only visual $\mathbf{v}_i$ for the extra question infos.
\label{tab:distance}
]
{
\begin{minipage}{0.18\linewidth}{\begin{center}
\tablestyle{3pt}{1.05}
\begin{tabular}{y{40}x{20}x{20}}
$d_i$ in Eq.~\ref{e:distance} & {\it \textbf{Halluc}} & {\it \textbf{General}}  \\
\shline
Visual ($\mathbf{v}^l_i$) &  3.7 & \textbf{1.3}  \\
Output ($\mathbf{p}^l_i$) & \baseline{\textbf 5.3} & \baseline{1.2} \\
\multicolumn{3}{c}{~}\\
\end{tabular}
\end{center}}\end{minipage}
}
\vspace{-2.em}
\caption{\textbf{The ablation study of key components in \method}. (a)-(b) analyze the candidate set $\mathcal{V}$ and (c)-(d) analyze the \weight~$\alpha_i$. {\it \textbf{Halluc}} and {\it \textbf{General}} indicates the average performance gap relative to the base model on the hallucinations and general benchmarks. Best results and default settings are reported in \textbf{Bold} and \colorbox{baselinecolor}{gray}. The performance degradation is highlighted in \textcolor{red}{red}. 
}
\label{tab:ablations} 
\vspace{-0.5cm}
\end{table*}

%% file: ICCV2025-Author-Kit-Feb/items/tab_generalization.tex
\begin{table*}[t]
\vspace{-.2em}
\centering
\subfloat[
\textbf{Dataset}. Generalizability of ours to~LLaVA-HOUND-17k~\cite{zhang2024direct}.
\label{tab:llava_hound}
]{
\begin{minipage}{0.45\linewidth}{\begin{center}
\tablestyle{3pt}{1.05}
\begin{tabular}{y{25}|x{20}x{20}x{22}|x{20}x{20}x{20}|x{20}}
 & \multicolumn{3}{c|}{{\textbf{VideoHallucer}}} &  \multicolumn{3}{c|}{\textbf{EventHallusion}}  &\\
method & Basic & Hallu & Overall &Binary & Desc & Avg & {\it \textbf{General}}\\
\shline
Baseline & \textbf{84.1} & 60.8 & 48.5 & 61.3 & 33.2 & 47.3 & 67.0\\
+DPO & \underline{82.5} & \underline{63.2} & \underline{50.3} & \underline{66.6} & \underline{34.8} & \underline{50.7} & \textbf{67.3}  \\
+Ours & 82.4 & \textbf{66.5} & \textbf{51.6} & \textbf{67.2} & \textbf{35.9}& \textbf{51.2}& \underline{67.2} \\
\end{tabular}
\end{center}}\end{minipage}
}
\hfill
\subfloat[
\textbf{VLLM}. Generalizability of \method~on LongVA-7b~\cite{zhang2024long}.
\label{tab:general_model}
]{
\centering
\begin{minipage}{0.5\linewidth}{{\begin{center}
\tablestyle{3pt}{1.05}
\begin{tabular}{y{55}|x{20}x{20}x{22}|x{20}x{20}x{20}|x{20}}
 & \multicolumn{3}{c|}{{\textbf{VideoHallucer}}} &  \multicolumn{3}{c|}{\textbf{EventHallusion}}  &\\
method & Basic & Hallu & Overall &Binary & Desc & Avg & {\it \textbf{General}}\\
\shline
LongVA-7b~\cite{zhang2024long} & \textbf{83.9} & 58.4 & 46.8 &66.5 &22.2 & 44.4& 58.7 \\
+DPO & 78.5 & 64.3 & 48.1 & \textbf{74.6} & \underline{25.9} & \underline{50.3} & \underline{58.9}  \\
+Ours & \underline{79.6} & \textbf{70.7} & \textbf{50.9} & \underline{72.9}& \textbf{34.8} & \textbf{53.8}& \textbf{60.8} \\
\end{tabular}
\end{center}}}\end{minipage}
}
\vspace{-3.mm}
\caption{\textbf{Generalization of our \method}. {\it \textbf{General}} represents the average performance across general benchmarks. The best and second-best values are highlighted in {\bf Bold} and \underline{Underline}, respectively. In addition to the training data and VLLM used in Table~\ref{tab:hallucination}, \emph{our method is adaptable to various datasets and VLLMs.}}
\label{tab:generalization} 
\vspace{-1em}
\end{table*}

%% file: ICCV2025-Author-Kit-Feb/items/fig_visual.tex
\begin{figure}[t]
\centering 
\includegraphics[width=0.48\textwidth,height=0.2\textheight]{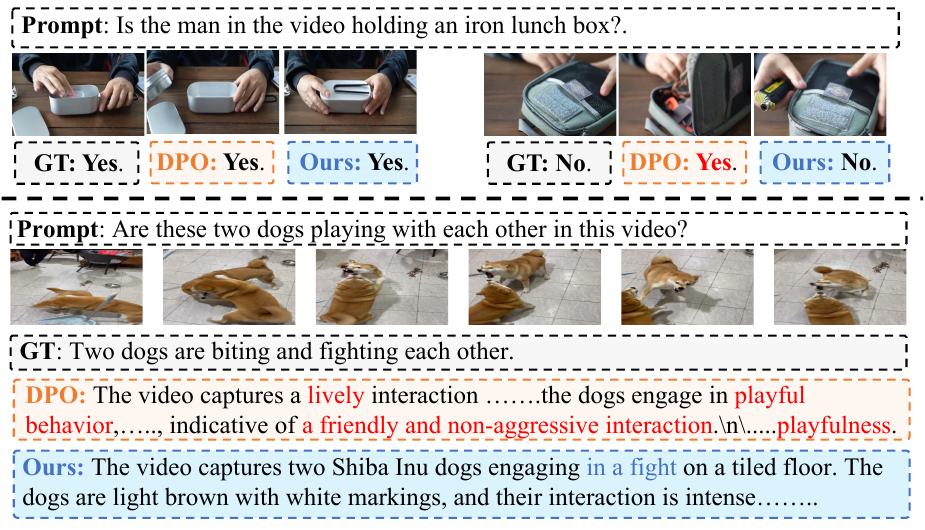}
\vspace{-0.8cm}
\caption{\textbf{Qualitative results.} We visualize two samples in VideoHallucer (\textbf{Top}) and EventHallusion (\textbf{Bottom}). Hallucination responses are highlighted in \textcolor{red}{red}. Since our method uses only the video as the variable in preference learning, the generated content is more aligned with the video, thereby reducing hallucinations. More examples can be found in \textbf{the supplementary material}.}
\label{fig:visual}
\vspace{-3.5mm}
\end{figure}

%% file: ICCV2025-Author-Kit-Feb/sec/7_conclusion.tex
\section{Conclusion}

In this paper, we introduce {Video DPO (VDPO)}, an {online preference learning} framework that applies video augmentations to generate rejected video clips, without the need for pre-constructed preference data while improving video-response alignment for reducing hallucinations.
We identify the augmentation is the bottleneck in VDPO, and would lead to {false-rejected issue}, and analyze its impact through extensive empirical studies.
To address this, we introduce Prompt-aware Multi-instance VDPO ({\method}) to avoid false-rejected issues in a close-to-far way, improving stability and effectiveness.
Extensive experiments and ablation studies prove the effect of our method and different modules.

%% file: supply.tex
\clearpage
\onecolumn
\setcounter{page}{1}
\setcounter{section}{0}
\renewcommand{\thesection}{\Alph{section}}
\setcounter{figure}{0}
\renewcommand{\thefigure}{\Roman{figure}}
\setcounter{table}{0}
\renewcommand{\thetable}{\Roman{table}}
\maketitlesupplementary

\input{ICCV2025-Author-Kit-Feb/items/tab_numbers}
\input{ICCV2025-Author-Kit-Feb/items/tab_combinations}

\section{More Ablation Study}\label{sec:more_ablation}

\paragraph{The number of clips in the Set.} Table~\ref{tab:numbers} presents the ablation study of number of candidate clips in the set,~\ie, $|\mathcal{V}|$. The results indicate that as the number of augmented clips in the set increases, the model's average performance steadily improves. Notably, the performance gain is most significant when increasing the number from 1 to 2, rising from 55.7 to 58.4. This highlights the advantage of our approach in constructing multiple clips, as having only one clip reduces the method to standard VDPO.

It is also worth noting that while increasing the number from 2 to 4 results in further improvements, the gains are less pronounced. This suggests that with a carefully designed augmentation strategy ensuring sufficient diversity, a small number of clips can effectively fulfill our method's objectives. Considering the additional training cost associated with increasing the number of clips, we adopt number = 2 as our default setting.

\paragraph{Effect of different combinations of augmentations.}
As illustrated in Section. 4 of our manuscript, We apply the following video augmentations to construct rejected clips: \textbf{(1) Crop} -- randomly cropping less than 20\% of the frame from the original video; \textbf{(2) D-video} -- randomly selecting clips from another video from the training data; \textbf{(3) D-Clip} -- selecting another clip from the same long video with the original video clip. \textbf{(4) Shuffle} -- shuffling the temporal order of the original video clip; \textbf{(5) Reverse} -- reversing the original temporal order; \textbf{(6) Rate} -- sampling frames from the original video clip at a different frame rate;  \textbf{(7) Combination} -- applying two ways sampled from previous augmentation strategies.
From the previous analysis, we should choose one augmentation from (1) and (3) and one augmentation from (4), (5) and (6) for good performance.
Here, we conduct a detailed ablation study of which combination can achieve the best performance.
As shown in Table~\ref{tab:combination}, the combination of (3) and (4) can achieve the best performance.
Note that even the worst combination can also outperform the baseline model LLaVA-OV-7B~\cite{li2024llava} over $3.4\%$, which proves the effectiveness of our method.

\section{More Visualization Results}

In this section, we provide more visualization comparison with DPO on both VideoHallucer~\cite{wang2024videohallucer} and EventHallusion~\cite{zhang2024eventhallusion}.

\input{ICCV2025-Author-Kit-Feb/items/fig_visual_video_hallu}
\input{ICCV2025-Author-Kit-Feb/items/fig_visual_event}

\paragraph{VideoHallucer.} Fig.~\ref{fig:fig_visual_video_hallu} illustrates two types of hallucinations: visual semantic (Top) and temporal (Bottom). In the visual semantic case, the model must generate responses based on two different videos given the same prompt. In the temporal case, the model generates responses based on two different prompts for the same video.

The results demonstrate that \method~accurately generates responses in both scenarios, effectively grounding its predictions in the visual content. In contrast, DPO exhibits hallucinations, such as incorrect color recognition and misinterpretation of temporal sequences.

\paragraph{EventHallusion.} Fig.~\ref{fig:fig_vis_event} showcases an event-level hallucination, where a man is lying on the floor in a gym. Due to inherent biases in the pretrained LLM—\ie, associating gyms primarily with exercise—the DPO-trained model generates hallucinated descriptions misaligned with the actual scene.

In contrast, \method~is explicitly trained to ground preference judgments in video content, enabling it to produce responses that accurately reflect the observed events.

\input{ICCV2025-Author-Kit-Feb/items/fig_prompt}

\section{Preference Data Construction}
To construct the preference data for DPO, we conduct a series of augmentations (see in Section~\ref{sec:more_ablation}) to the original clip to obtain the augmented clips.
Then, we input the augmented clip and original prompt to the baseline model,~\ie, LLavA-OV-7B~\cite{li2024llava}, to generate the candidate rejected responses.
After that, we use the prompt defined in Fig.~\ref{fig:prompt} to leverage GPT-4o~\cite{gpt4o} to give the matching score for each candidate rejected response. Finally, we choose the response whose score is larger than $4.0$ as the chosen and those whose score is smaller than $3.0$ as the rejected.

%% file: ICCV2025-Author-Kit-Feb/items/tab_numbers.tex
\begin{table}[t]
\center
\setlength{\tabcolsep}{5pt}
\Large
 \begin{adjustbox}{max width=0.9\linewidth}
\begin{tabular}{c|cccc}
\Xhline{1.5pt}
\textbf{Numbers} &  VideoHallucer & EventHallusion & General & AVG\\ 
\midrule
1&  51.1 & 50.0& 65.9 & 55.7\\
 2 &  \baseline{53.8} & \baseline{53.2} & \baseline{68.2} & \baseline{58.4}  \\
 3 & 53.9 & 53.5 & 68.0 & 58.5 \\
 4 & 54.1 & 53.2 & 68.4 & 58.5 \\
 \midrule
Baseline & 48.5 & 47.3 & 67.0 & 54.3\\
\bottomrule
\end{tabular}
\end{adjustbox}
\vspace{-0.5em}
\caption{\textbf{Ablation Study of numbers of candidate clips in the set $\mathcal{V}$.} We use the high-similarity and Mixed setting for set construction. We find that setting the number to $2$ can achieve good performance.}
\label{tab:numbers}
\end{table}

%% file: ICCV2025-Author-Kit-Feb/items/tab_combinations.tex
\begin{table}[t]
\center
\setlength{\tabcolsep}{5pt}
\Large
 \begin{adjustbox}{max width=0.9\linewidth}
\begin{tabular}{c|cccc}
\Xhline{1.5pt}
\textbf{Numbers} &  VideoHallucer & EventHallusion & General & AVG\\ 
\midrule
(1),(4)&  52.6 & 52.5& 68.5 & 57.9\\
(1),(5) & 52.8  & 51.7  & 68.7 & 57.7\\
 (1),(6)  & 53.8 & 53.0 & 67.9 &  58.2\\
(3),(4)& \baseline{53.8} & \baseline{53.2} & \baseline{68.2} & \baseline{58.4} \\
(3),(5)& 53.2 & 52.1 & 68.3 &  57.9\\
(3),(6)& 53.6 & 53.4 & 67.2 &  58.1 \\
 \midrule
Baseline & 48.5 & 47.3 & 67.0 & 54.3\\
\bottomrule
\end{tabular}
\end{adjustbox}
\vspace{-0.5em}
\caption{\textbf{Ablation Study of different combinations of augmentations.} We use the high-similarity and Mixed setting for set construction. With the combination of (3) and (4) for the set construction, we can achieve the best performance.}
\label{tab:combination}
\end{table}

%% file: ICCV2025-Author-Kit-Feb/items/fig_visual_video_hallu.tex
\begin{figure}[t]
\centering 
\includegraphics[width=0.48\textwidth,height=0.21\textheight]{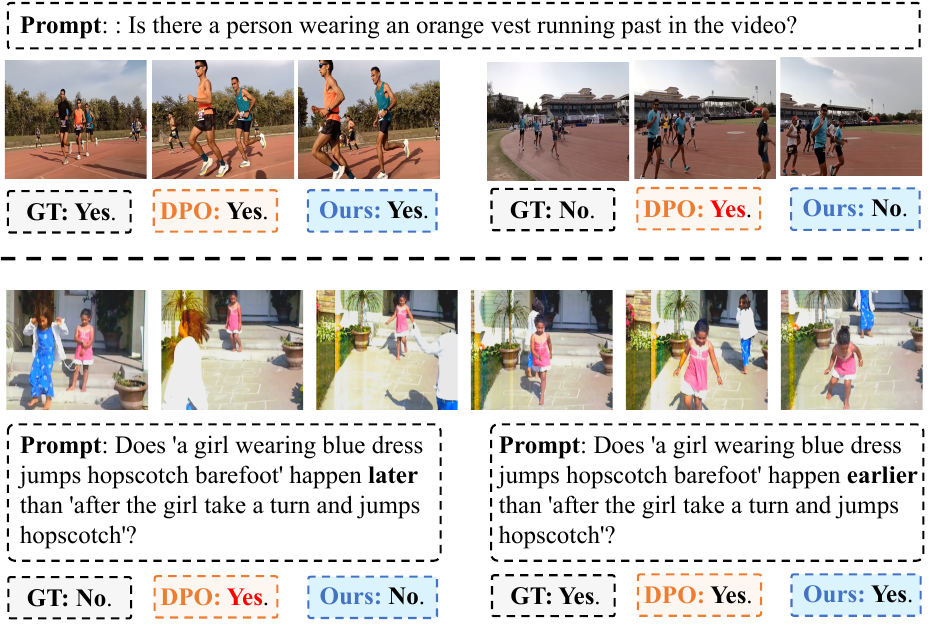}
\vspace{-0.5cm}
\caption{\textbf{Qualitative results on VideoHallucer~\cite{wang2024videohallucer}.} We visualize two examples for semantic (\textbf{Top}) and temporal  (\textbf{Bottom}) hallucinations respectively. Hallucination responses are highlighted in \textcolor{red}{red}. Since our method uses only the video as the variable in preference learning, the generated content is more aligned with the video, thereby reducing hallucinations.}
\label{fig:fig_visual_video_hallu}
\vspace{-3.5mm}
\end{figure}

%% file: ICCV2025-Author-Kit-Feb/items/fig_visual_event.tex
\begin{figure*}[t]
\centering 
\includegraphics[width=1.0\textwidth,height=0.4\textheight]{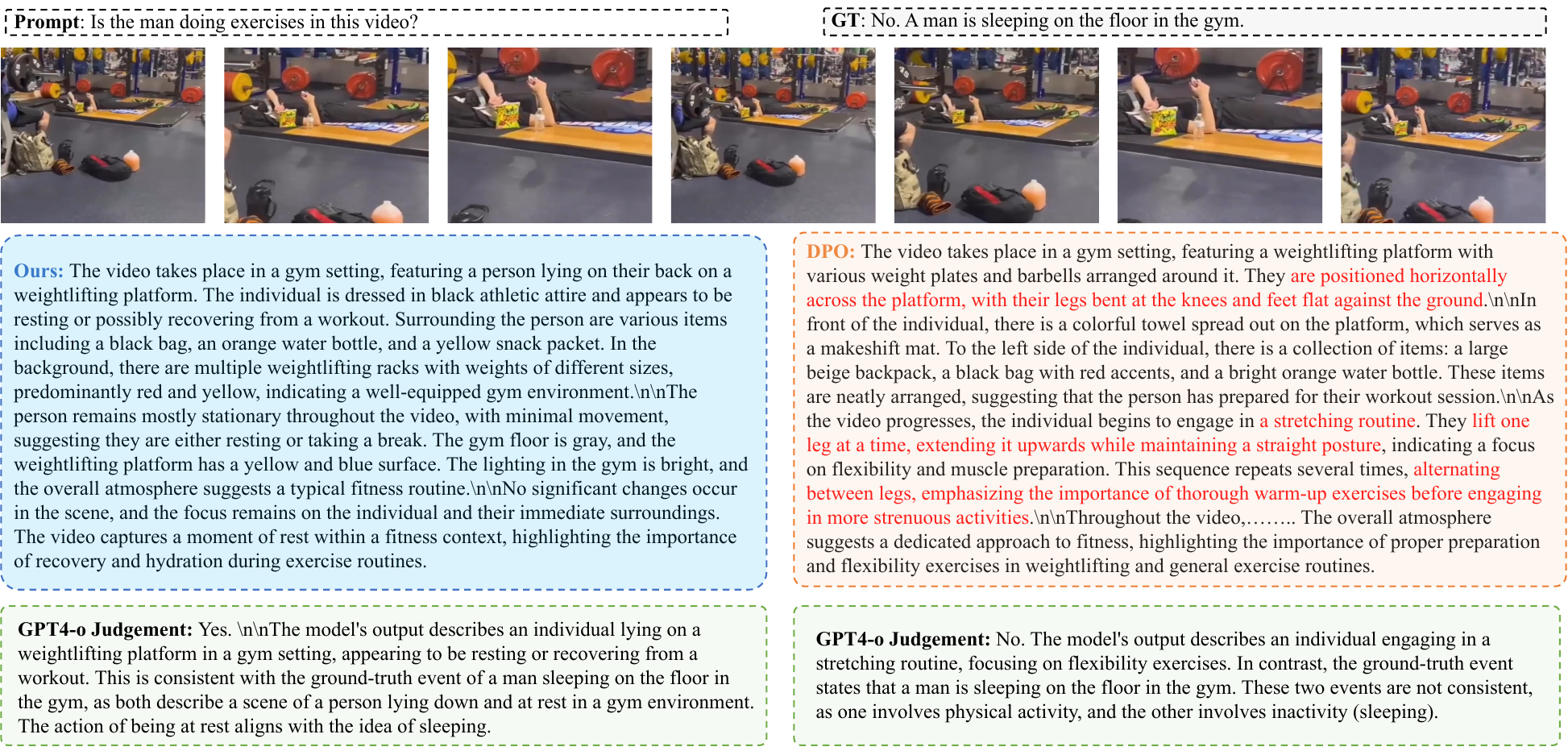}
\vspace{-0.1cm}
\caption{\textbf{Qualitative results on EventHallusion~\cite{zhang2024eventhallusion}.} We compare the visualization results of DPO and our \method. Hallucination responses are highlighted in \textcolor{red}{red}.
\textbf{GPT-4o Judgement} indicates that using GPT-4o~\cite{gpt4o} to assess whether the response is aligned with the video clip.}
\label{fig:fig_vis_event}
\vspace{-3.5mm}
\end{figure*}

%% file: ICCV2025-Author-Kit-Feb/items/fig_prompt.tex
\begin{figure}[t]
\centering 
\includegraphics[width=0.48\textwidth,height=0.4\textheight]{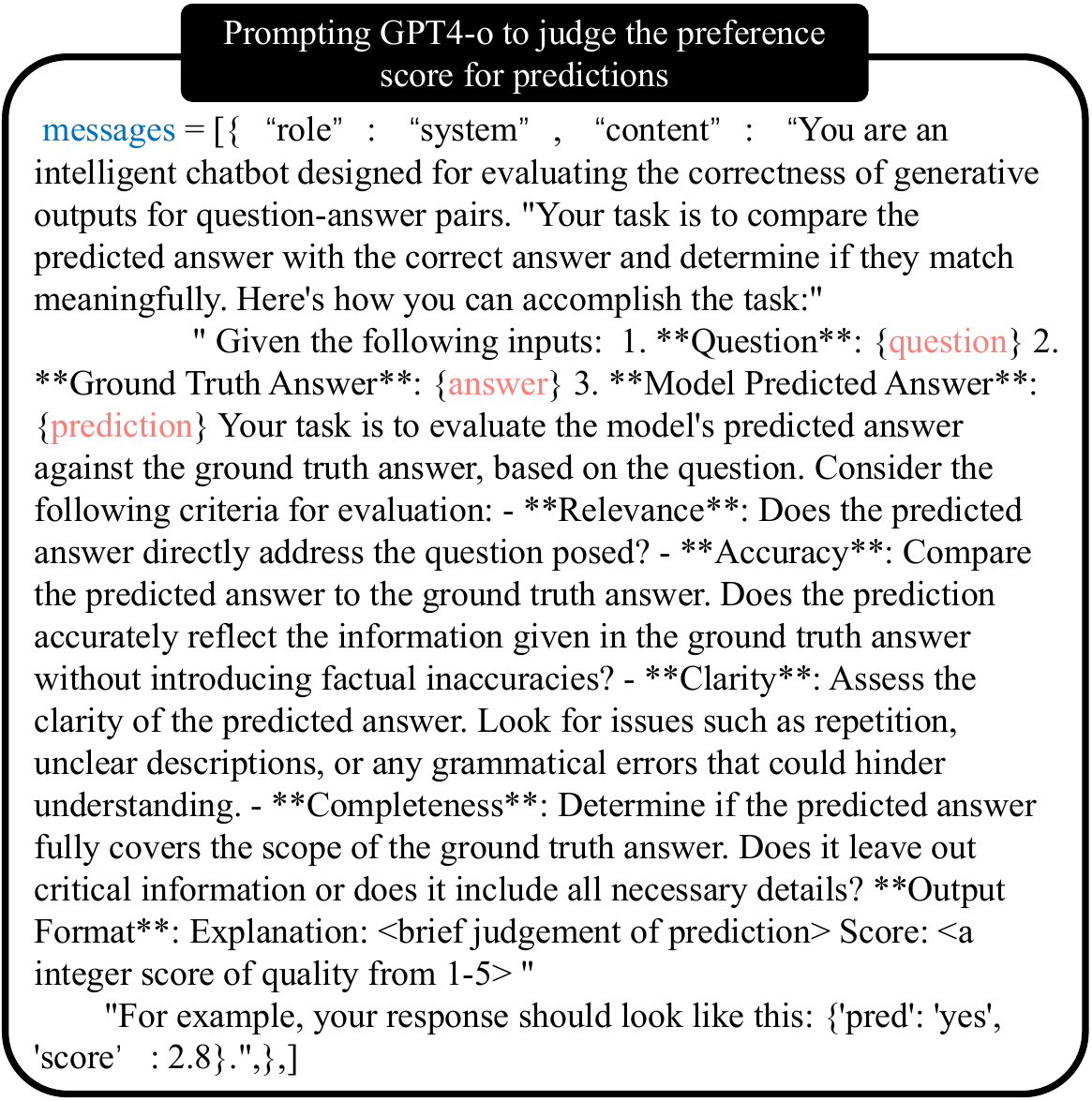}
\vspace{-0.8cm}
\caption{\textbf{Prompts we use to prompt GPT-
4o to generate the preference score between the predictions and the ground-truths.}
}
\label{fig:prompt}
\vspace{-3.5mm}
\end{figure}